\documentclass[10pt,twocolumn,letterpaper]{article}

\usepackage{cvpr}
\usepackage{times}

\usepackage{amsmath}
\usepackage{amssymb}

\usepackage{xcolor,colortbl}

\usepackage{algorithm2e}
\usepackage{multirow}

\usepackage[normalem]{ulem} 

\usepackage[pagebackref=true,breaklinks=true,letterpaper=true,colorlinks,bookmarks=false]{hyperref}

\cvprfinalcopy 

\ifcvprfinal\fi

\title
      {Rephrasing visual questions
      by specifying the entropy of the answer distribution} 

\author{%
 Kento Terao, 
 Toru Tamaki,\\
 Bisser Raytchev, 
 Kazufumi Kaneda\\
 Hiroshima University, Japan
 \and
 Shun'ichi Satoh\\
 National Institute of Informatics, Japan
}

\usepackage{cite}

\usepackage[pdftex]{graphicx}
\graphicspath{{./images/}}

\usepackage{amsmath}

\usepackage{multirow}

\newcommand{\Prt}[2][0]{
\ifnum #1 = 0 \hspace{-0.3mm}\left(\hspace{-0.1mm} #2 \hspace{-0.1mm}\right)\hspace{-0.1mm} \fi
\ifnum #1 = 1 (#2) \fi
\ifnum #1 = 2 \bigl(#2\bigr) \fi
\ifnum #1 = 3 \Bigl(#2\Bigr) \fi
\ifnum #1 = 4 \biggl(#2\biggr) \fi
\ifnum #1 = 5 \Biggl(#2\Biggr) \fi
}

\def\unk{unk }

\begin{document}

\maketitle

\begin{abstract}
Visual question answering (VQA) is a task of answering a visual question that is a pair of question and image. Some visual questions are ambiguous and some are clear, and it may be appropriate to change the ambiguity of questions from situation to situation. However, this issue has not been addressed by any prior work.
We propose a novel task, rephrasing the questions
by controlling the ambiguity of the questions.
The ambiguity of a visual question is defined
by the use of the entropy of the answer distribution predicted by a VQA model.
The proposed model rephrases a source question given with an image
so that the rephrased question has the ambiguity (or entropy) specified by users.
We propose two learning strategies
to train the proposed model with the VQA v2 dataset,
which has no ambiguity information.
We demonstrate the advantage of our approach that can control the ambiguity of the rephrased questions,
and an interesting observation that it is harder to increase than to reduce ambiguity.
\end{abstract}

\section{Introduction}

\begin{figure}[t]
    \centering

    \includegraphics[width=\linewidth]{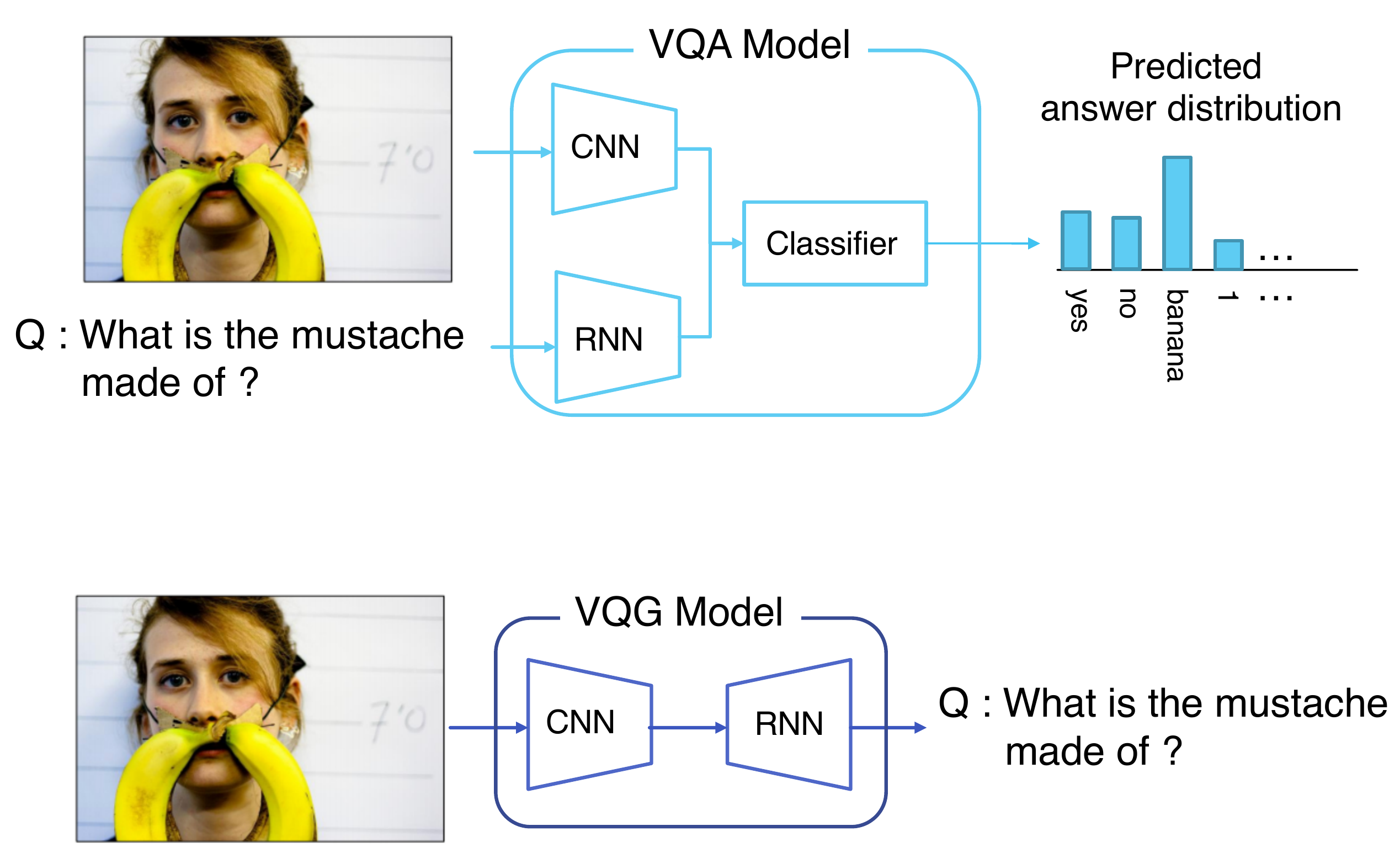}

    \caption{
    Two related tasks on question answering.
    (top) Visual Question Answering (VQA) takes 
    a visual question (a pair of image and question text) as input
    and answers the question by predicting the answer distributions of multiple answer categories.
    (bottom) Visual Question Generation (VQG) takes an image as input and generates
    a question about the image.
    }

    \label{fig:vqa_and_vqg}
\end{figure}

Visual Question Answering (VQA) is one of the most challenging tasks in computer vision
\cite{WU2017CVIU,VQAv1}:
given a pair of question text and image (a visual question),
a system is asked to answer the question.
There is also a related task called Visual Question Generation (VQG),
which generates questions about the given image (Figure \ref{fig:vqa_and_vqg}).
These tasks have been attracting a lot of attention in recent years
because it has a large potential to impact many applications
such as smart support for the visually impaired \cite{Gurari_2018_CVPR},
providing instructions to autonomous robots \cite{embodiedqa},
and intelligent interaction between humans and machines \cite{visual_dialog}.

\begin{figure}[t]
    \centering

    \includegraphics[width=\linewidth]{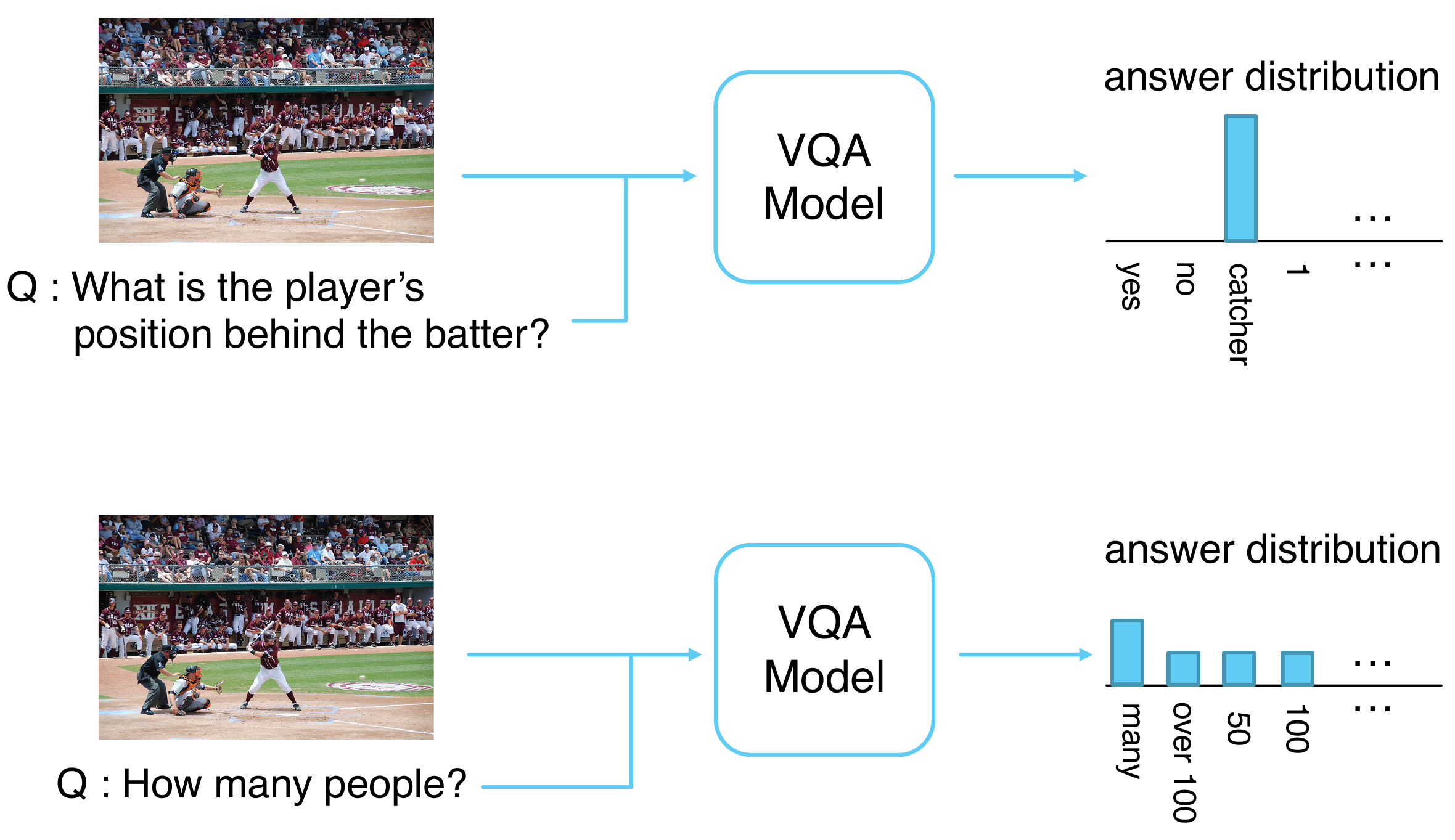}

    \medskip

    \includegraphics[width=\linewidth]{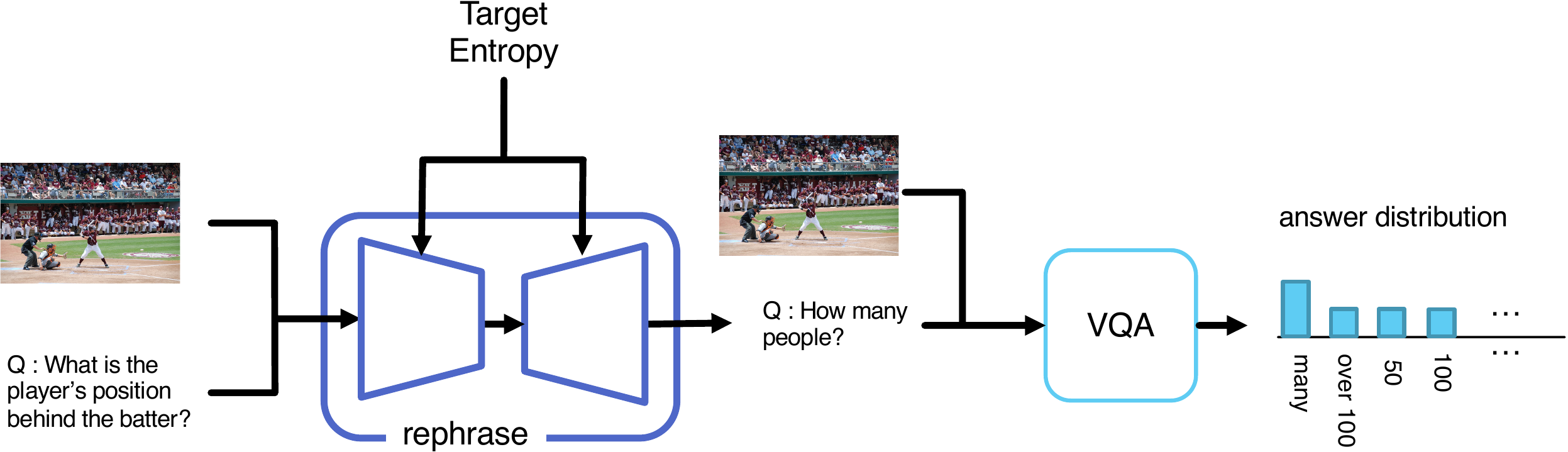}
    
    \caption{Two different types of questions with different degrees of ambiguity.
    (top) A unique answer is expected to a clear question.
    (middle) Multiple equally possible answers are predicted because of the question's ambiguity.
    (bottom) Our model is able to rephrase one question into another with different ambiguity.
    }

    \label{fig:easy_difficult_questions}
\end{figure}

There may be two different scenarios when asking a visual question:
one needs a clear question and the other requires an ambiguous question (Figure \ref{fig:easy_difficult_questions}).
The former scenario involves the case of a smart support system for visually impaired people \cite{Gurari_2018_CVPR}
who are asking the system about the image.
The context of such a visual question should be as clear as possible so that the system can provide a specific answer to the question.
But sometimes users' questions may be vague, and 
the system returns multiple answers that are equally possible.
This can be a source of confusion, because the user is not certain whether the answers are correct.
The latter scenario happens when you want to ask a visual question that might start a conversation or ask opinions.
It is good for such visual questions to be unclear,
so that different responses can be expected \cite{mostafazadeh2016ACL_VQG}.
This kind of ambiguity is due to various factors \cite{Bhattacharya_2019_ICCV},
such as the subjectivity of questions (\ie, asking about opinions on the image)
or the difficulty of the task (\ie, counting a number of objects).
Such scenarios lead to the need to modify (rephrase, or control) the ambiguity of visual questions.
Prior work on VQG either generates questions from answers \cite{Shah_2019_CVPR}
or categories of answers \cite{krishna2019information},
or learns VQA and VQG simultaneously \cite{Li_2018_CVPR,Shah_2019_CVPR}.
However, to the best of our knowledge,
no work has handled the ambiguity when generating or rephrasing questions.

In this paper, we propose a model that modifies a given visual question
in such a way that the rephrased visual question has the ambiguity specified by the user
(bottom of Figure \ref{fig:easy_difficult_questions}).
We define the ambiguity of a visual question
with the entropy value of the answer distribution
predicted by a VQA model.
Let $A$ be the set of answers, and the entropy $H(A)$ is defined by
\begin{equation}
H(A) = -\sum_{a \in A} P(a) \ln P(a).
\end{equation}
In general, entropy is large when the distribution is broad, and small when it has a narrow peak.
If the entropy is large, we think that the visual question is ambiguous because it results in various answers,
and if the entropy is small, it is a clear visual question that provides a unique answer (or fewer possible answers).
Our proposed model is closely related to a common VQG model that takes an input image $I$ and generates a question text $Q_G$.
The difference is that our model also takes
a \emph{source question} $Q_S$ and a \emph{target entropy} value $E_T$ as input.
The question $Q_G$ generated by our model is then fed into a VQA model to predict an answer distribution,
from which the entropy $E_G$ is computed.
The proposed model learns to minimize the error between the specified target entropy $E_T$
and the entropy $E_G$ (as shown in Figure \ref{fig:vqg_overview}) .
In addition, the loss between the generated question $Q_G$ and a target question $Q_T$,
which is used as a reference text, is also minimized in order to impose restrictions
on the generated question to be similar to questions in the dataset.
We also propose two learning strategies to train our model.
Our task is novel and hence there are no datasets for this task so far,
therefore we utilize an existing VQA dataset, the VQA v2 \cite{balanced_vqa_v2},
to train our rephrasing model.

\section{Related Work}

\subsection{VQG}

Our work is closely related to VQG, which is
proposed by Mostafazadeh \etal \cite{mostafazadeh2016ACL_VQG}.
VQG can be done by using image captioning models,
however there is a gap from questions 
generated by humans even if the evaluation score (such as BLEU) is high.
Therefore by focusing on the fact 
that humans use words related to abstract concepts when asking questions,
they proposed the VQG task
and created a VQG dataset containing more abstract words than image captioning datasets.
Their objective is to generate a natural question 
to help start a conversation about a given image.
This is similar to generating a more ambiguous question with higher entropy,
but their model is not able to control the question ambiguity.

Other works tried to generate multiple different questions.
Zhang \etal \cite{DBLP:conf/ijcai/ZhangQYYZ17} proposed a model that can
generate visually grounded questions with diverse types for a single image.
In this model, not only images but also captions generated
by a dense captioning model are used as input.
This model can generate multiple types of questions that refer to different image regions described by dense captioning.
Fan \etal \cite{DBLP:conf/ijcai/FanWLLH18} 
used a conditional variational autoencoder
to generate  multiple questions from each question type and a given image.
They use two score functions
that take into account probabilities of generating the target question
and the question type, and generate top $k$ questions.
Krishna \etal \cite{krishna2019information}
built a model that maximizes mutual information between the image,
the expected answer, and the generated question,
based on the observation that
VQG should be a task aimed at extracting concepts
of a particular category from images.
They used a variational autoencoder with two latent spaces;
one is between the image and the expected answer, 
and the other is between the image and the answer category.
Questions can be generated by specifying the answer category.
These works can generate diverse questions about the image,
but the ambiguity of the generated questions has not been taken into account.

\subsection{Joint learning of VQG and VQA}

Some works use VQG for improving VQA performance.
Li \etal \cite{Li_2018_CVPR} proposed
a model that simultaneously learns VQA and VQG,
and showed that the joint learning
based on the MUTAN \cite{MUTAN} model improves
the performance of VQA.
Shah \etal \cite{Shah_2019_CVPR} proposed a framework
that uses the cycle consistency between VQA and VQG models
to make VQA  more robust to linguistic variations in visual questions.
In this framework, 
VQG generates a visual question based on the answer predicted by VQA, 
and VQA predicts the answer to the generated question again.

Our work also uses VQA and VQG as components of the proposed model,
however our goal is not to improve the performance of VQA, 
instead we use VQA to evaluate the entropy of rephrased questions
as a proxy of ambiguity.

\subsection{Paraphrasing}

Our question rephrasing is also similar to paraphrasing,
which expresses the same meaning in different phrases.
Liu \etal \cite{Liu_2019_ICCV} proposed
to modify (or paraphrase) a generated caption text
so that modified captions have more
diverse expressions like in human-provided captions.
They used a score function to evaluate the syntactic complexity
to select more descriptive caption candidates.
However the syntax of question texts does not reflect
the ambiguity of visual questions.

\section{Proposed model}

\begin{figure*}[t]
    \centering
    \hfill    
    \begin{minipage}{.4\linewidth}
    \centering
    \includegraphics[width=\linewidth]{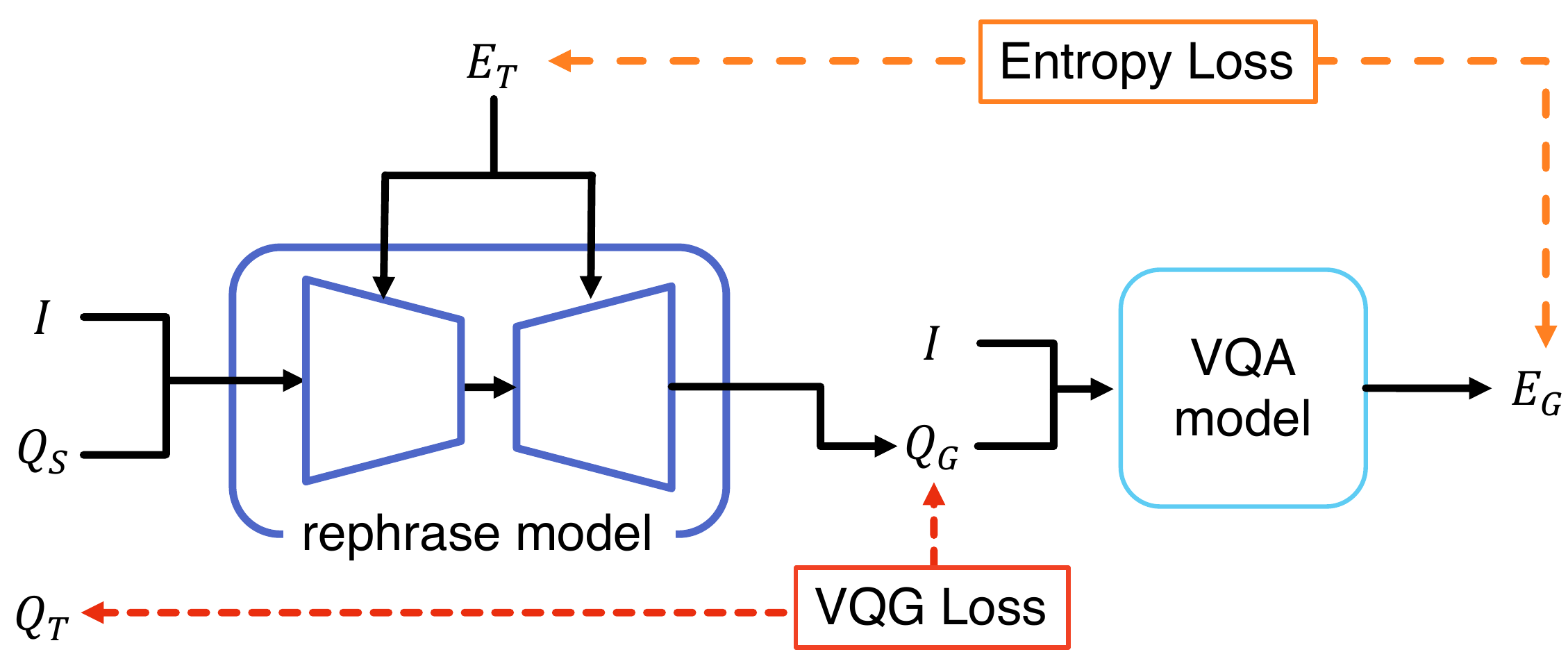}
    \end{minipage}
    \hfill
    \begin{minipage}{.45\linewidth}
    \centering
    \includegraphics[width=\linewidth]{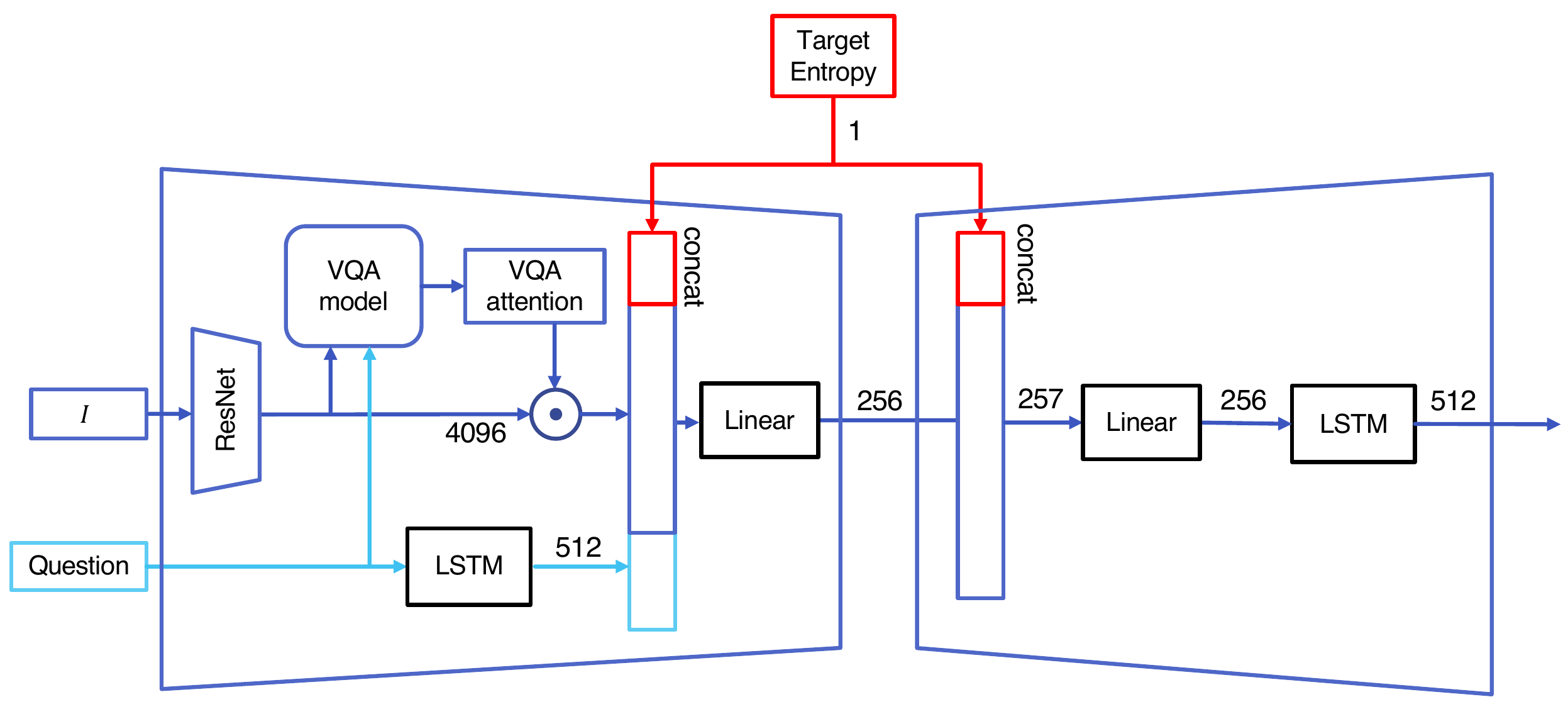}    
    \end{minipage}
    \hfil
    
    \caption{
    (Left) Overview of the proposed model.
    $Q_s, Q_t, Q_g$ are source, target, and generated
    questions.
    $E_s, E_t, E_g$ are entropy values of predicted
    answer distributions by the VQA model to questions 
    $Q_s, Q_t, Q_g$, respectively.
    $I$ is the image.
    (Right) Details of the encoder and decoder.
    }

    \label{fig:vqg_overview}
    \label{fig:encoder_decoder}
    
\end{figure*}

The proposed model is shown in Figure \ref{fig:vqg_overview} (left).
Given source visual question $Q_{S}$, image $I$, and target entropy $E_{T}$,
the rephrase model outputs generated (or rephrased) 
question $Q_{G}$ whose entropy $E_G$ is close to $E_{T}$.
Entropy $E_G$ is calculated from the predicted answer distribution of a pre-trained VQA model,
which is frozen during the training of the the rephrasing model.

\subsection{Model Details}

The rephrase model has an encoder-decoder structure (Figure \ref{fig:encoder_decoder} (right)).

\subsubsection{Encoder}

Unlike VQG models which only take an image as input,
our rephrase encoder takes image $I$, source question $Q_S$, and target entropy $E_T$
as input.
A pre-trained ResNet152 \cite{He_2016_CVPR}
is used to extract image features from $I$,
and an LSTM is used for extracting text information from the source question $Q_S$.
Target entropy $E_T$, which is a scalar, is also used
as input to specify the desired ambiguity of the rephrased question.
These are concatenated and passed to a linear layer.

Furthermore, we use a pre-trained VQA model inside the encoder
to modify the image features.
This is because we want the model to rephrase the source question,
rather than generating an unrelated new text.
To this end, we utilize the attention that the VQA model is used.
VQA with attention mechanisms is known to work well \cite{Anderson_2018_CVPR}
because it is an important key to correctly answer to the question
by knowing which part of the image is relevant.
Therefore the VQA attention is expected to encode
the part used in the source question.
By attending to that part and emphasizing the image features with it,
the hidden feature passed from the encoder to the decoder
is expected to retain information about where is the important part in the image
relevant for the source question.

\subsubsection{Decoder}

The decoder receives the output of the encoder with the attention 
information, and is expected to generate a question that is similar to the source question
in terms of the attention, which is what we want to achieve, that is, rephrasing.
The decoder also takes target entropy $E_T$, as in the encoder,
to enforce again the model to take the target entropy into account.
These are concatenated and passed to a linear layer, followed by an LSTM
to generate a rephrased question, which is represented by a word sequence 
$Q_{G0}, \ldots, Q_{Gn}$.

\subsubsection{Entropy computation by VQA}

The rephrase model with the encoder-decoder architecture is
enough for rephrasing source questions, but we need to check
if the rephrased question has the ambiguity specified by target entropy.
Therefore we add the VQA model, which is same as the one used in the encoder.

The rephrased (generated by the decoder) question $Q_G$  
and the same image $I$ are then
passed to the VQA model for predicting the answer distribution.
Entropy $E_G$ is computed from this answer distribution,
and expected to express the ambiguity of $Q_G$.

\subsection{Training}

There is no dataset for this task, and it is difficult to collect such datasets
because the ambiguity of questions is subjective and hard to define.
Instead, we propose to use an existing VQA dataset with the following protocol.

\subsubsection{Losses}

We use two losses.
One is an entropy loss $L_\mathrm{Ent}$,
which is the error between the entropy $E_G$ of the generated question
and the target entropy $E_T$;
\begin{equation}
  L_\mathrm{Ent} = (E_T - E_G)^2.
\end{equation}
This forces the model to generate questions with the specified target entropy value.

The other loss is a VQG loss $L_\mathrm{VQG}$ \cite{Shah_2019_CVPR,krishna2019information},
which is
the sum of the negative log likelihood over the generated word sequence $Q_{G0}, \ldots, Q_{Gn}$,
given the reference (target) word sequence $Q_{T0}, \ldots, Q_{Tn}$ as follows:
\begin{equation}
  L_\mathrm{VQG} = 
  - \frac{1}{n}
  \sum_{i=1}^n 
  \log p(Q_{Gi} \ | \ Q_{Ti}).
\end{equation}

Putting them together, the final loss for our model is
$L = L_\mathrm{VQG} + \lambda L_\mathrm{Ent}$,
where $\lambda$ is a hyper parameter.

\subsubsection{Learning strategies}

We introduce two different learning strategies: (1) noise and (2) sampling.

\paragraph{Setting 1: Noise}

We use the source question $Q_S$ as the target question $Q_T$.
Let $E_S$ be the source entropy, that is, the entropy value computed from
the answer distribution of the VQA model with the image $I$ and the source question $Q_S$.
The target entropy $E_T$ is defined by
\begin{equation}
    E_T = E_S + \epsilon, \ \epsilon \sim U(-1, 1)
\end{equation}
where noise $\epsilon$ is subject to a uniform noise $U(-1, 1)$ between $-1$ and $+1$.

The motivation for this setting is to enforce the model
to generate a question $Q_G$ similar to the source question $Q_S$ (via the VQG loss),
but having different entropy value (via the entropy loss).
Therefore the model is expected to behave like a rephraser.

\paragraph{Setting 2: Sampling}

Many VQA datasets have a set of multiple questions $Q_I$ for the same image $I$.
Therefore, we use one of the different questions
that belong to the same image $I$ to which the source question $Q_S$ belongs.
The target question $Q_T$ is randomly sampled from $Q_I$, excluding $Q_S$, that is,
\begin{equation}
    Q_T \sim Q_I / Q_S.
\end{equation}
Target entropy $E_T$ is computed by using $Q_T$ with the VQA model.

In this setting the generated question $Q_G$ may not be a rephrased version
of the source question $Q_S$ because the sampled target question $Q_T$
might not be asking similar concept with $Q_S$.
However the entropy $E_G$ is expected to be similar to the specified target entropy $E_T$,
because $E_T$ is computed from the actual question sentence $Q_T$.
In contrast, the noise setting uses the same question for $Q_S$ and $Q_T$
but asks the model to change the entropy, which might confuse the model.

We compare these two settings in the experiments.

\section{Experiments} 

\subsection{Datasets and setting}
We use VQA v2 \cite{balanced_vqa_v2},
a standard benchmark dataset for VQA
and used in annual challenges since 2016\footnote{\url{https://visualqa.org/}}.
It consists of training, validation, and test sets.
To train the proposed model, we use the training set
(82,783 images, 443,757 questions, and 4,437,570 answers).
We use the validation set
(40,504 images, 214,354 questions, and 2,143,540 answers)
for rephrasing visual questions and analysis.

We use Pythia v0.1 \cite{pythia2018arxiv,singh2018pythia}
as a pre-trained VQA model for the encoder and for entropy computation.
As in prior work \cite{Anderson_2018_CVPR,lu2019vilbert,Teney_2018_CVPR,Yu_2019_CVPR},
3129 answers in the training set
that occur at least 8 times are chosen as candidates,
which results in a multi-class problem
predicting answer distributions of 3129 dimension.

\subsubsection{Implementation Details}

The LSTMs used in the encoder and decoder have one layer with hidden size of 512.
The batch size was 64 and the maximum iterations were set to 44000.
Adam with a learning rate of 0.0005 was used for the optimization.
To back-propagate the losses over the generated questions,
we used the Gumbel-Softmax trick 
\cite{DBLP:conf/iclr/JangGP17}
with a temperature parameter $\tau = 0.01$.

\subsubsection{Evaluation metrics}
We use the error between the target entropy $E_T$ and the generated entropy $E_G$
in order to evaluate whether the proposed model generates a question
with the specified ambiguity.

In addition,
we use standard similarity metrics 
(BLEU-4, CIDEr, METEOR, ROUGE-L \cite{sharma2017nlgeval}),
commonly used for captioning and VQG,
to evaluate the quality of the generated questions.

We also use the number of unique questions to evaluate the diversity of the generated questions.

\subsubsection{Comparisons}

We compare the following three different settings for each of the two learning strategies.

\begin{itemize}

    \item \textbf{Noise/Sampling Pretrain}:
    The model learns only with the VQG loss without using the entropy loss.
    This is equivalent to setting $\lambda=0$.

    \item \textbf{Noise/Sampling}:
    The model learns from scratch
    using both the VQG loss and entropy loss with an equal weight $\lambda=1$.
    
    \item \textbf{Noise/Sampling-FT}:
    The model learns via fine-tuning (FT)
    using both the VQG loss and entropy loss with an equal weight $\lambda=1$.
    The pre-training with the VQG loss only (\ie, Pretrain setting)
    is used for a warm start.

\end{itemize}

\subsection{Rephrasing}

When rephrasing source questions,
our model requires the target entropy $E_T$.
For evaluating rephrasing results, we vary $E_T$ with respect to $E_S$.
More specifically,
given a source question $Q_S$, we compute source entropy $E_S$,
and then add a scalar $\Delta$ to $E_S$; $E_T = E_S + \Delta$.
Here $\Delta$ is varied from $-2$ to $+2$ by the step of 0.5;
\ie, $\Delta \in \{ -2.0, -1.5, -1.0, -0.5, 0.0, +0.5, +1.0, +1.5, +2.0 \}$.
Note that entropy should be positive,
hence we exclude questions when $E_S + \Delta < 0$.
This means that 
there are less valid questions 
for large negative $\Delta$.
The number of questions and unique questions (diversity, see below)
for each value of $\Delta$ is shown in
Table \ref{tab:statistics_diff}.

\begin{table}[t]
    \centering
    \caption{Number of questions in the VQA v2 validation set for each $\Delta$.}
    \label{tab:statistics_diff}
    \begin{tabular}{cccc}
$\Delta$  & $E_T$         & \# questions & Diversity \\ 
\hline
$-2.0$    & $E_S - 2.0$   & 47956        & 23964     \\ 
$-1.5$    & $E_S - 1.5$   & 63875        & 30330     \\ 
$-1.0$    & $E_S - 1.0$   & 82629        & 36621     \\ 
$-0.5$    & $E_S - 0.5$   & 128235       & 59821     \\ 
$\pm 0.0$ & $E_S \pm 0.0$ & 214354       & 81565     \\ 
$+2.0$    & $E_S + 0.5$   & 214354       & 81565     \\ 
$+1.5$    & $E_S + 1.0$   & 214354       & 81565     \\ 
$+1.0$    & $E_S + 1.5$   & 214354       & 81565     \\ 
$+0.5$    & $E_S + 2.0$   & 214354       & 81565     \\ 
\hline
    \end{tabular}
\end{table}

Table \ref{tab:specify_entropy_difference} shows quantitative results for
each value of $\Delta$ for six different situations.
The column $|E_T - E_G|$ shows
averages and standard deviations (stds)
of errors between the specified target entropy $E_T$
and the entropy $E_G$ of the rephrased question.
This should be small so that rephrased questions $Q_G$ have the desired ambiguity
with the specified entropy.

Columns BLUE-4, CIDEr, METEOR, and ROUGE-L
are similarity scores between source and rephrased questions.
Unlike captioning or VQG tasks where generated texts should be close to those in the training set,
higher scores do not mean better results in this task.
This is because we rephrase (modify) the source question
and generate a different question with different ambiguity,
therefore these scores are not necessary high.
In other words, if these scores are too high,
then it means that the model doesn't rephrase (or modify) the source question.

The column Diversity shows the number of unique questions generated by the model.
The original VQA v2 validation set has at most
81565 different unique questions (the column Diversty of Table \ref{tab:statistics_diff}).
The diversity score often decreases when the variety of questions generated by the model
is limited (\ie, the same questions are often generated).

In Noise Pretrain and Sampling Pretrain settings, 
the model was trained without the entropy loss
and can not control the generated entropy $E_G$. Therefore
their errors $E_T - E_G$ are consistently high.
The model is trained so that the rephrased questions should be identical to
the source questions, then similarity scores are quite high, particularly for Noise Pretrain.

In the Noise and Sampling setting, the model was trained from scratch without pretraining,
therefore the training was not stable.
Entropy errors are smaller than those of Pretrain settings,
however similarity scores and diversity are extremely low.
A possible reason might be that reasonable questions cannot be generated
at the initial stage of learning, while the model tries to reduce the entropy loss.

In Noise-FT and Sampling-FT settings, the model was fine-tuned after the pretraining.
This achieves often the smallest entropy error for each of the $\Delta$ values,
and similarity and diversity scores are higher than those without pretraining.

The Noise-FT model can generate a variety of questions (high diversity),
but has problems generating questions when increasing entropy (positive $\Delta$ values).
On the other hand, the Sampling-FT model can control the entropy, 
but the variety of questions that can be generated is small (low diversity).

\begin{figure}[t]
    \centering
    \includegraphics[width=\linewidth]{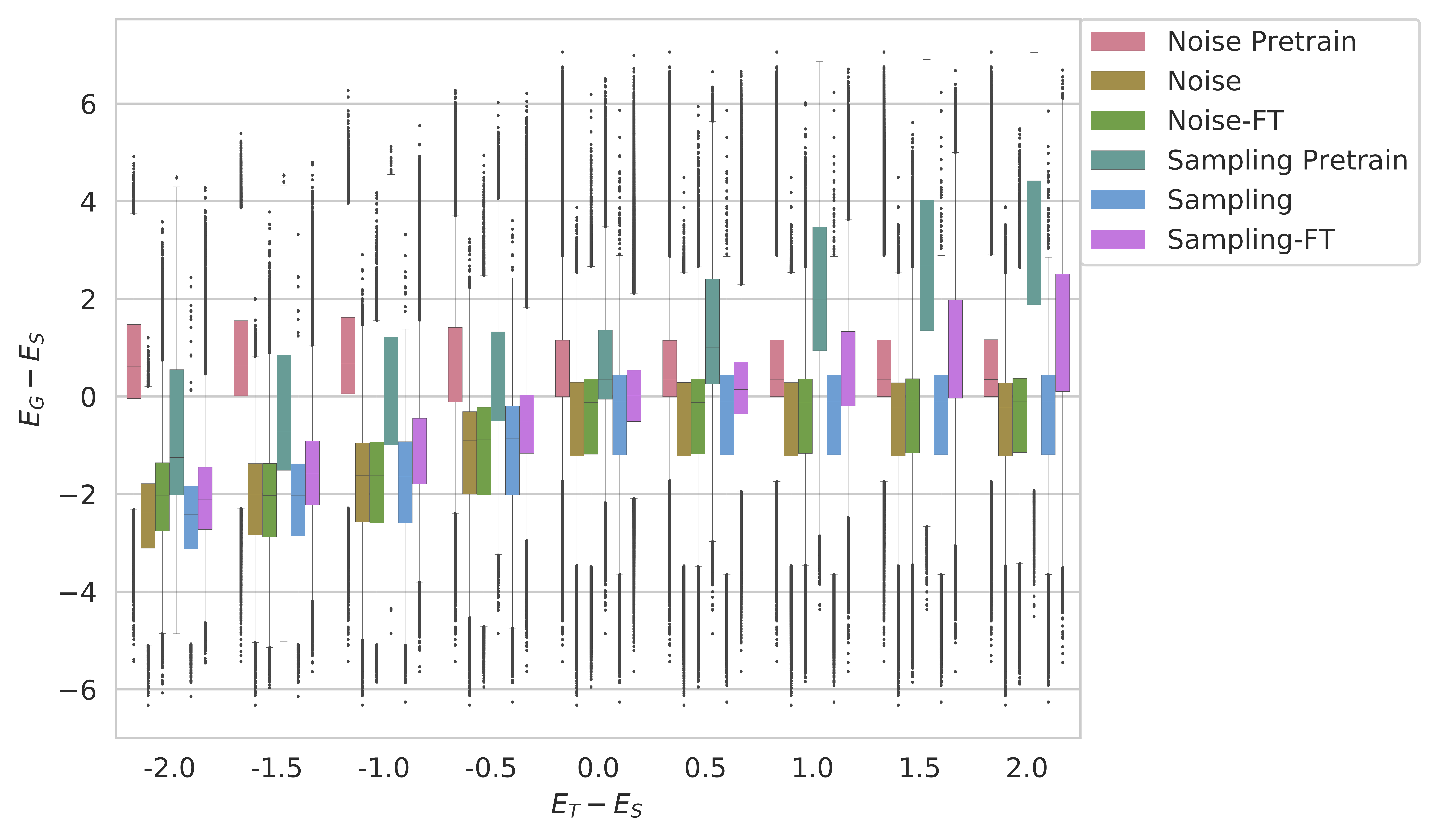}
    \caption{
    Box plot of entropy errors for each value of $\Delta$ over all questions.
    The horizontal axis is the specified value of $\Delta = E_T - E_S$,
    and the vertical axis is the entropy error $E_G - E_T$ (without taking absolute).
    }
    \label{fig:boxplot}
\end{figure}

We further investigate how the model controls entropy errors
by using box plots shown in Figure \ref{fig:boxplot}.
The horizontal axis ($E_T-E_S$) corresponds to each $\Delta$ value in Table \ref{tab:specify_entropy_difference}
from $-2.0$ to $+2.0$.
The vertical axis is signed entropy difference $E_G - E_S$
while the entropy error column in
Table \ref{tab:specify_entropy_difference} shows absolute errors $|E_T -E_G|$.
Ideally, this plot should be linear with a unit slope
because $E_G$ and $E_T$ are the same in the ideal case.
We can see that Noise Pretrain is almost flat, which means 
the model has no control on the entropy.
In contrast, Sampling Pretrain has too much large slope
and it fails to control entropy too. 
The best-looking result is Sampling-FT;
it has almost a unit slop,
and $E_G - E_S$ is almost the same with $E_T - E_S$.
For the other three cases, it looks difficult to make $E_G-E_S$ larger
when $E_T-E_S$ is positive.
Thus it turns out that increasing entropy (more ambiguous) is harder
than decreasing entropy (less ambiguous).

\begin{figure}[t]
\centering

\begin{minipage}{.2\linewidth}
\includegraphics[width=\linewidth]{VQG/example/image_1.jpg}
\end{minipage}
\scalebox{0.4}{
\begin{tabular}{lccp{16em}cp{6em}r}
            &   & $E_S$ & $Q_S$ \\ \hline
            &   & 2.899 & What is the word written on the field ? \\
                                                                 \\            
            & $\Delta$  & $E_T$ & $Q_G$                                                      & $E_G$ & VQA top-1     & $E_G - E_T$  \\
\hline
Noise       & $-1$  & 0.899 & what is the word on the net ?                              & 4.601 & polo          & 3.702  \\
Pretrain    & $0$   & 2.899 & what is the word on the net ?                              & 4.601 & polo          & 1.702  \\
            & $+1$  & 4.899 & what is the word on the jersey ?                           & 4.414 & yes           & -0.485 \\
\hline
Noise-FT    & $-1$  & 0.899 & the stripe on the stripe stripe stripe ?                   & 2.314 & yes           & 1.415  \\
            & $0$   & 2.899 & does the people on the stripe stripe ?                     & 0.452 & no            & -2.447 \\
            & $+1$  & 4.899 & game is this ?                                             & 0.572 & no            & -4.327 \\
\hline
Sampling    & $-1$  & 0.899 & what color is the ball ?                                   & 5.020 & white         & 4.121  \\
Pretrain    & $0$   & 2.899 & what is the man doing ?                                    & 5.392 & hit ball      & 2.493  \\
            & $+1$  & 4.899 & what is the name of the company advertised on the banner ? & 1.326 & polo          & -3.573 \\
\hline
Sampling-FT & $-1$  & 0.899 & is this a professional baseball game ?                     & 0.702 & no            & -0.197 \\
            & $0$   & 2.899 & who is going downhill ?                                    & 2.671 & tennis player & -0.228 \\
            & $+1$  & 4.899 & who is going on the bike ?                                 & 4.756 & man           & -0.143
\end{tabular}
}

\vspace{.5cm}

\begin{minipage}{.2\linewidth}
\includegraphics[width=\linewidth]{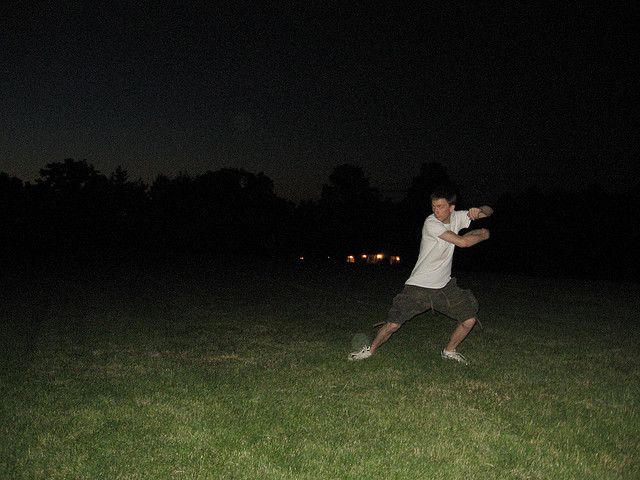}
\end{minipage}
\scalebox{0.4}{
\begin{tabular}{lccp{16em}cp{6em}r}
            &   & $E_S$ & $Q_S$ \\ \hline
            &   & 2.161 & What is he doing at night ? \\
                                                                 \\            
            & $\Delta$  & $E_T$ & $Q_G$                                                      & $E_G$ & VQA top-1     & $E_G - E_T$  \\
\hline
Noise       & $-1$  & 0.161             & what is he doing at this game ?                        & 3.851 & throwing & 3.691  \\
Pretrain    & $0$   & 2.161             & what is he doing at this game ?                        & 3.851 & throwing & 1.691  \\
            & $+1$  & 4.161             & what is he doing at this game ?                        & 3.851 & throwing & -0.309 \\
\hline
Noise-FT    & $-1$  & 0.161             & would the people have fun ? sit on the stripe stripe ? & 0.279 & no       & 0.119  \\
            & $0$   & 2.161             & is this an elderly person ?                            & 0.023 & no       & -2.137 \\
            & $+1$  & 4.161             & is this a young ball ?                                 & 0.275 & no       & -3.885 \\
\hline
Sampling    & $-1$  & 0.161             & is the man wearing a hat ?                             & 0.126 & no       & -0.035 \\
Pretrain    & $0$   & 2.161             & what is the man doing ?                                & 5.181 & nothing  & 3.02   \\
            & $+1$  & 4.161             & what is the man doing ?                                & 5.181 & nothing  & 1.02   \\
\hline
Sampling-FT & $-1$  & 0.161             & will the kite in the picture ?                         & 0.601 & no       & 0.441  \\
            & $0$   & 2.161             & which two people going downhill ?                      & 2.341 & man      & 0.181  \\
            & $+1$  & 4.161             & why does the person have two feet in the air ?         & 4.09  & playing frisbee & -0.07
\end{tabular}
}

\vspace{.5cm}

\begin{minipage}{.2\linewidth}
\includegraphics[width=\linewidth]{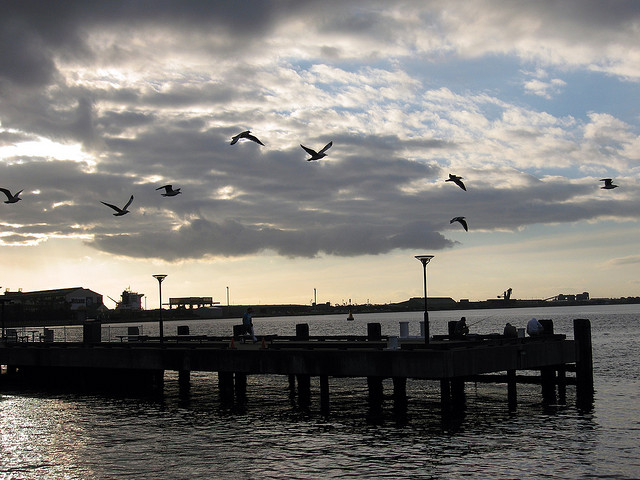}
\end{minipage}
\scalebox{0.4}{
\begin{tabular}{lccp{16em}cp{6em}r}
            &   & $E_S$ & $Q_S$ \\ \hline
            &   & 2.773 & \parbox{15em}{What is traveling under the bridge archway ?} \\
                                                                 \\            
            & $\Delta$  & $E_T$ & $Q_G$                                                      & $E_G$ & VQA top-1     & $E_G - E_T$  \\
\hline
Noise       & $-1$  & 0.773             & what is parked at the ramp ?                                                         & 3.457 & no     & 2.684  \\
Pretrain    & $0$   & 2.773             & what is parked at the ramp ?                                                         & 3.457 & no     & 0.684  \\
            & $+1$  & 4.773             & what is parked at the ramp ?                                                         & 3.457 & no     & -1.316 \\
\hline
Noise-FT    & $-1$  & 0.773             & the picture stripe wet ?                                                             & 2.528 & no     & 1.755  \\
            & $0$   & 2.773             & does the people have dinner stripe ?                                                 & 0.586 & no     & -2.187 \\
            & $+1$  & 4.773             & does the people stripe \unk stripe in the stripe next stripe ? & 0.312 & no     & -4.462 \\
\hline
Sampling    & $-1$  & 0.773             & is the boat in the water ?                                                           & 0.049 & no     & -0.724 \\
Pretrain    & $0$   & 2.773             & what is the weather like ?                                                           & 4.522 & wet    & 1.749  \\
            & $+1$  & 4.773             & where is the bird ?                                                                  & 3.123 & no     & -1.651 \\
\hline
Sampling-FT & $-1$  & 0.773             & is this a passenger train ?                                                          & 0.126 & no     & 0.648  \\
            & $0$   & 2.773             & who is going on the truck ?                                                          & 1.62  & birds  & -1.153 \\
            & $+1$  & 4.773             & who is going on the car ?                                                            & 2.229 & birds  & -2.545
\end{tabular}
}

    \caption{Examples of rephrasing. There are three visual questions (images $I$ are on the left,
    and source questions $Q_S$ are at the top of each table.
    Top-1 answers predicted by VQA are shown for reference.
    }
    
    \label{fig:example}
\end{figure}

Figure \ref{fig:example} shows rephrasing results for three visual questions
with $\Delta \in \{-1, 0, 1\}$ in four situations.
In Noise Pretrain the model tends to generate the same question
even if the target entropy is changed, 
indicating that the model cannot control the entropy.
In Noise-FT and Sampling Pretrain, the model generates different questions
but entropy errors are large.
In contrast, the model in Sampling-FT has smaller entropy errors
with different and diverse questions.
A common issue to all situations is that rephrased questions $Q_G$ are sometimes
unrelated to the given images.
Our future work includes investigating
how to use the VQA attention more explicitly in the decoder to fix this issue.

\begin{table}[t]
 \centering
 \caption{Entropy errors, similarity and diversity scores of rephrasing results.}
 \label{tab:specify_entropy_difference}
 
\scalebox{0.5}{
 \begin{tabular}{c|l|cccccc}
$\Delta = E_T-E_S$    &         & $|E_T - E_G|$  & BLEU-4 & CIDEr  & METEOR & ROUGE-L & Diversity \\  
\hline
\multirow{6}{*}{\centering $-2.0$}
            & Noise Pretrain    & 2.69$\pm$1.20  & 69.91  & 691.17 & 46.97  & 87.54   & 26488     \\  
            & Noise             & 0.90$\pm$0.71  & 0.00   & 0.31   & 4.77   & 9.77    & 91        \\  
            & Noise-FT          & 0.91$\pm$0.77  & 0.19   & 1.95   & 6.08   & 21.72   & 8937      \\  
            & Sampling Pretrain & 1.46$\pm$1.35  & 8.73   & 57.19  & 15.61  & 45.47   & 2229      \\  
            & Sampling          & 0.82$\pm$0.66  & 0.14   & 1.12   & 5.58   & 24.91   & 183       \\  
            & Sampling-FT       & 0.91$\pm$0.83  & 0.74   & 5.38   & 6.87   & 27.96   & 1087      \\  
\hline
\multirow{6}{*}{\centering $-1.5$}
            & Noise Pretrain    & 2.31$\pm$1.18  & 71.13  & 701.98 & 47.80  & 88.07   & 33719     \\  
            & Noise             & 1.00$\pm$0.80  & 0.00   & 0.34   & 4.72   & 9.66    & 100       \\  
            & Noise-FT          & 0.97$\pm$0.81  & 0.27   & 3.38   & 7.45   & 24.13   & 9029      \\  
            & Sampling Pretrain & 1.40$\pm$1.27  & 9.56   & 61.63  & 15.75  & 45.55   & 2361      \\  
            & Sampling          & 0.95$\pm$0.77  & 0.15   & 1.35   & 5.60   & 25.04   & 214       \\  
            & Sampling-FT       & 0.91$\pm$0.82  & 0.84   & 5.59   & 6.92   & 28.00   & 1217      \\  
\hline
\multirow{6}{*}{\centering $-1.0$}
            & Noise Pretrain    & 1.94$\pm$1.17  & 71.95  & 709.68 & 48.41  & 88.45   & 40974     \\  
            & Noise             & 1.12$\pm$0.91  & 0.00   & 0.35   & 4.70   & 9.62    & 100       \\  
            & Noise-FT          & 1.08$\pm$0.92  & 0.30   & 3.35   & 7.42   & 24.02   & 10635     \\  
            & Sampling Pretrain & 1.38$\pm$1.22  & 11.19  & 72.19  & 16.37  & 46.51   & 2524      \\  
            & Sampling          & 1.07$\pm$0.89  & 0.20   & 1.61   & 5.63   & 25.13   & 254       \\  
            & Sampling-FT       & 0.93$\pm$0.83  & 0.80   & 5.46   & 6.87   & 27.73   & 1274      \\  
\hline
\multirow{6}{*}{\centering $-0.5$}
            & Noise Pretrain    & 1.36$\pm$1.17  & 70.49  & 696.61 & 47.46  & 87.74   & 66290     \\  
            & Noise             & 1.10$\pm$0.99  & 0.00   & 0.34   & 4.73   & 9.49    & 114       \\  
            & Noise-FT          & 1.07$\pm$0.99  & 0.37   & 3.56   & 7.53   & 24.66   & 13634     \\  
            & Sampling Pretrain & 1.18$\pm$1.16  & 12.14  & 79.20  & 16.86  & 47.36   & 3117      \\  
            & Sampling          & 1.08$\pm$0.96  & 0.25   & 1.81   & 5.78   & 25.75   & 345       \\  
            & Sampling-FT       & 0.89$\pm$0.88  & 1.06   & 6.52   & 7.02   & 28.10   & 1550      \\  
\hline
\multirow{6}{*}{\centering $\pm 0.0$}
            & Noise Pretrain    & 0.92$\pm$1.07  & 70.58  & 697.59 & 47.45  & 87.80   & 96075     \\  
            & Noise             & 1.03$\pm$1.02  & 0.00   & 0.31   & 4.72   & 9.19    & 124       \\  
            & Noise-FT          & 0.98$\pm$1.02  & 0.45   & 3.78   & 7.76   & 25.37   & 18449     \\  
            & Sampling Pretrain & 0.98$\pm$1.09  & 12.88  & 86.64  & 17.25  & 48.14   & 3734      \\  
            & Sampling          & 1.01$\pm$0.99  & 0.30   & 2.16   & 6.03   & 26.54   & 440       \\  
            & Sampling-FT       & 0.84$\pm$0.91  & 1.39   & 7.50   & 7.30   & 28.79   & 1846      \\  
\hline
\multirow{6}{*}{\centering $+0.5$}
            & Noise Pretrain    & 0.86$\pm$0.90  & 70.65  & 698.35 & 47.50  & 87.84   & 96193     \\  
            & Noise             & 1.26$\pm$1.13  & 0.00   & 0.31   & 4.73   & 9.19    & 114       \\  
            & Noise-FT          & 1.17$\pm$1.18  & 0.46   & 3.76   & 7.80   & 25.35   & 18251     \\  
            & Sampling Pretrain & 1.26$\pm$1.20  & 11.00  & 71.25  & 16.27  & 45.85   & 3312      \\  
            & Sampling          & 1.17$\pm$1.16  & 0.31   & 2.17   & 6.03   & 26.56   & 476       \\  
            & Sampling-FT       & 0.95$\pm$0.86  & 1.37   & 7.25   & 7.59   & 29.62   & 1846      \\  
\hline
\multirow{6}{*}{\centering $+1.0$}
            & Noise Pretrain    & 1.02$\pm$0.76  & 70.59  & 697.43 & 47.43  & 87.78   & 96210     \\  
            & Noise             & 1.63$\pm$1.22  & 0.00   & 0.31   & 4.73   & 9.20    & 110       \\  
            & Noise-FT          & 1.57$\pm$1.24  & 0.47   & 3.73   & 7.83   & 25.30   & 18334     \\  
            & Sampling Pretrain & 1.59$\pm$1.24  & 9.95   & 62.26  & 15.67  & 44.56   & 1874      \\  
            & Sampling          & 1.53$\pm$1.28  & 0.31   & 2.18   & 6.04   & 26.59   & 480       \\  
            & Sampling-FT       & 1.18$\pm$0.82  & 1.29   & 6.65   & 7.82   & 30.15   & 1712      \\  
\hline
\multirow{6}{*}{\centering $+1.5$}
            & Noise Pretrain    & 1.28$\pm$0.73  & 70.60  & 697.61 & 47.44  & 87.78   & 96358     \\  
            & Noise             & 2.05$\pm$1.30  & 0.00   & 0.31   & 4.73   & 9.20    & 118       \\  
            & Noise-FT          & 2.03$\pm$1.26  & 0.47   & 3.68   & 7.84   & 25.23   & 18373     \\  
            & Sampling Pretrain & 1.68$\pm$1.16  & 8.77   & 54.53  & 15.24  & 43.72   & 1853      \\  
            & Sampling          & 1.99$\pm$1.32  & 0.31   & 2.19   & 6.05   & 26.62   & 475       \\  
            & Sampling-FT       & 1.36$\pm$0.85  & 1.19   & 6.03   & 7.86   & 30.22   & 1501      \\  
\hline
\multirow{6}{*}{\centering $+2.0$}
            & Noise Pretrain    & 1.62$\pm$0.79  & 70.53  & 696.47 & 47.36  & 87.73   & 96312     \\  
            & Noise             & 2.52$\pm$1.35  & 0.00   & 0.31   & 4.73   & 9.20    & 114       \\  
            & Noise-FT          & 2.50$\pm$1.29  & 0.46   & 3.65   & 7.86   & 25.13   & 18376     \\  
            & Sampling Pretrain & 1.70$\pm$1.06  & 7.70   & 48.45  & 14.91  & 43.05   & 1718      \\  
            & Sampling          & 2.49$\pm$1.32  & 0.31   & 2.19   & 6.05   & 26.64   & 493       \\  
            & Sampling-FT       & 1.46$\pm$0.94  & 1.00   & 5.28   & 7.67   & 29.85   & 1345      \\  
\hline
\end{tabular}
}
\end{table}

\subsection{Impact of VQA attention}

\begin{figure}[t]
    \centering
    \includegraphics[width=\linewidth]{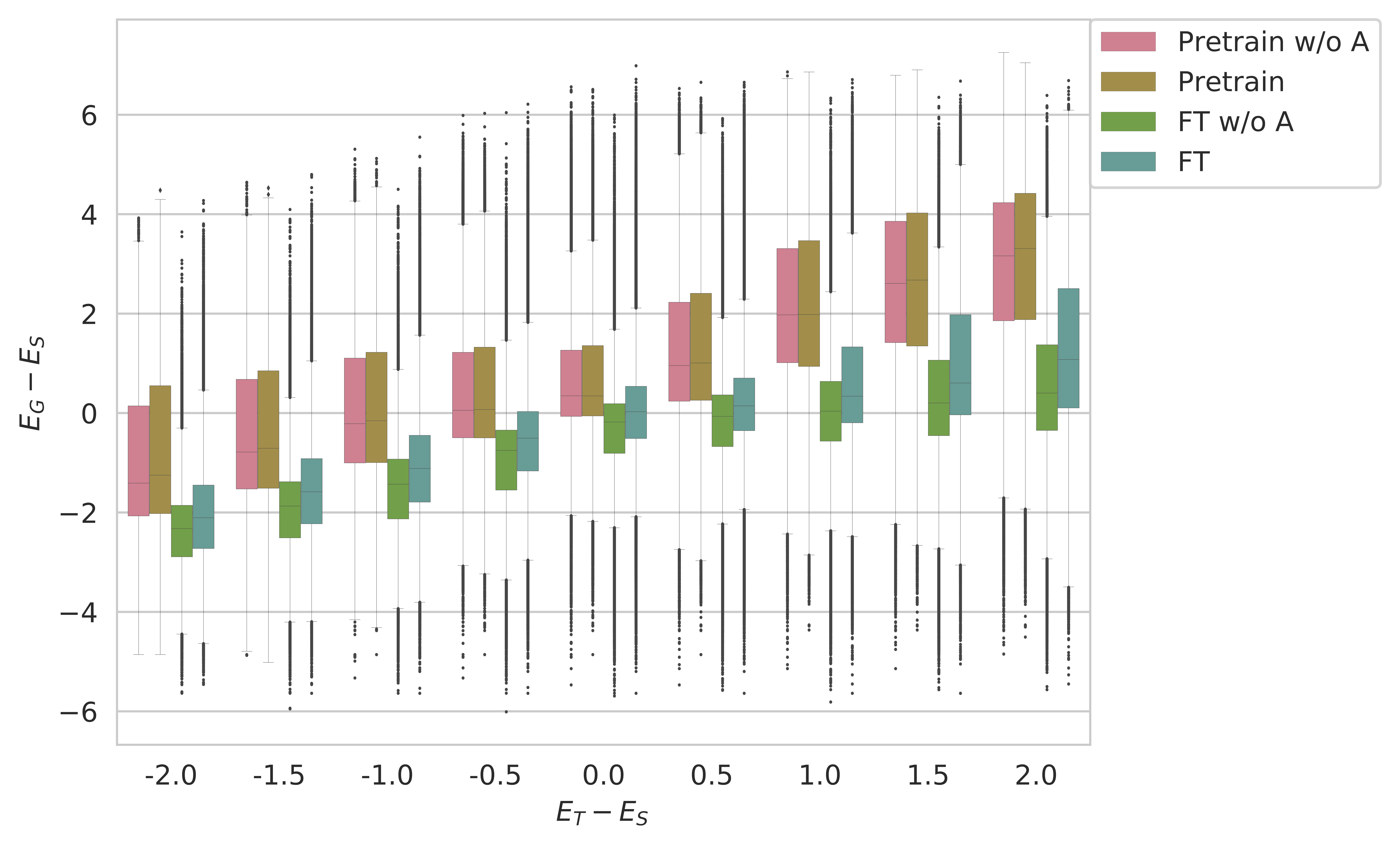}
    \caption{
    Effect of VQA attention on entropy errors for each value of $\Delta$ over all questions.
    Boxes with ``w/o A'' stands for Sampling Pretrain or FT without attention, otherwise with attention.
    The horizontal axis is the specified value of $\Delta = E_T - E_S$,
    and the vertical axis is the entropy error $E_G - E_T$ (without taking absolute).
     }
    \label{fig:boxplot_att}
\end{figure}

For two situations, Sampling-Pretrain and Sampling-FT,
we investigate the effect of the presence or absence of the VQA attention
on rephrasing questions.
When the model does not use the VQA attention, image features of ResNet152 are directly used.
Results are shown in Table \ref{tab:attention_difference}.
When VQA attention was used, all scores except Diversity increased.
In Sampling Pretrain entropy errors increased when attention was used,
while in Sampling-FT, entropy errors decreased when $\Delta > 0$.
We can see this effect in the box plot of Figure \ref{fig:boxplot_att}.
For Sampling Pretrain, results are similar whether VQA attention is used or not,
but for Sampling-FT, results with attention are clearly improved compared to those without attention.

The reason why diversity increased when attention was not used,
is that
the model can pay attention to various parts of the image.
If the VQA attention is used, the model restricts the image features to the important parts,
and as a result of this the variety of the generated questions becomes limited.

\begin{figure}[t]
\centering

\begin{minipage}{.2\linewidth}
\includegraphics[width=\linewidth]{VQG/example/image_1.jpg}
\end{minipage}
\scalebox{0.4}{
\begin{tabular}{lccp{16em}cp{6em}r}
            &   & $E_S$ & $Q_S$ \\ \hline
            &   & 2.899 & What is the word written on the field ? \\
                                                                 \\            
Sampling & $\Delta$  & $E_T$ & $Q_G$                                               & $E_G$ & VQA top-1     & $E_G - E_T$  \\
\hline
Pretrain & $-1$  & 0.899 & what is the man holding ?                                  & 4.066 & nothing       & 3.167  \\
w/o A    & $0$   & 2.899 & what is the man doing ?                                    & 5.392 & hit ball      & 2.493  \\
         & $+1$  & 4.899 & what is the man holding in his left hand ?                 & 3.978 & serve         & -0.921 \\
\hline
Pretrain & $-1$  & 0.899 & what color is the ball ?                                   & 5.020 & white         & 4.121  \\
         & $0$   & 2.899 & what is the man doing ?                                    & 5.392 & hit ball      & 2.493  \\
         & $+1$  & 4.899 & what is the name of the company advertised on the banner ? & 1.326 & polo          & -3.573 \\
\hline
FT w/o A & $-1$  & 0.899 & is there a lot facing to turn ?                            & 0.415 & no            & -0.485 \\
         & $0$   & 2.899 & is there a lot facing to turn to turn ?                    & 0.325 & yes           & -2.574 \\
         & $+1$  & 4.899 & is there a lot facing to turn turn ?                       & 0.385 & yes           & -4.514 \\
\hline
FT       & $-1$  & 0.899 & is this a professional baseball game ?                     & 0.702 & no            & -0.197 \\
         & $0$   & 2.899 & who is going downhill ?                                    & 2.671 & tennis player & -0.228 \\
         & $+1$  & 4.899 & who is going on the bike ?                                 & 4.756 & man           & -0.143
\end{tabular}
}

\vspace{.5cm}

\begin{minipage}{.2\linewidth}
\includegraphics[width=\linewidth]{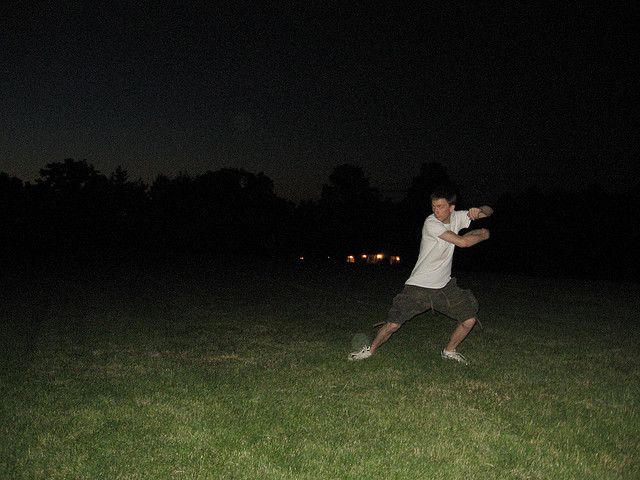}
\end{minipage}
\scalebox{0.4}{
\begin{tabular}{lccp{16em}cp{6em}r}
            &   & $E_S$ & $Q_S$ \\ \hline
            &   & 2.161 & What is he doing at night ? \\
                                                      \\            
Sampling & $\Delta$  & $E_T$ & $Q_G$                                               & $E_G$ & VQA top-1     & $E_G - E_T$  \\
\hline
Pretrain & $-1$  & 0.161 & is the man wearing a shirt ?                            & 0.452 & no      & 0.291  \\
w/o A    & $0$   & 2.161 & what is the man doing ?                                 & 5.181 & nothing & 3.02   \\
         & $+1$  & 4.161 & what is the man standing on ?                           & 4.898 & nothing & 0.737  \\
\hline
Pretrain & $-1$  & 0.161 & is the man wearing a hat ?                              & 0.126 & no      & -0.035 \\
         & $0$   & 2.161 & what is the man doing ?                                 & 5.181 & nothing & 3.02   \\
         & $+1$  & 4.161 & what is the man doing ?                                 & 5.181 & nothing & 1.02   \\
\hline
FT w/o A & $-1$  & 0.161 & is there a stop light ?                                 & 0.069 & no      & -0.091 \\
         & $0$   & 2.161 & is there a stop light ?                                 & 0.057 & no      & -2.103 \\
         & $+1$  & 4.161 & where do turn turn turn turn turn turn turn turn turn ? & 1.638 & left    & -2.522 \\
\hline
FT       & $-1$  & 0.161 & will the kite in the picture ?                          & 0.601 & no      & 0.441  \\
         & $0$   & 2.161 & which two people going downhill ?                       & 2.341 & man     & 0.181  \\
         & $+1$  & 4.161 & why does the person have two feet in the air ?          & 4.090 & playing frisbee & -0.07  \\
\end{tabular}
}

\vspace{.5cm}

\begin{minipage}{.2\linewidth}
\includegraphics[width=\linewidth]{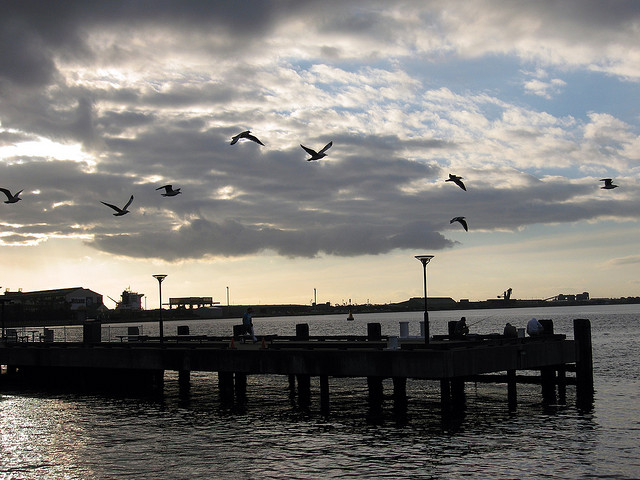}
\end{minipage}
\scalebox{0.4}{
\begin{tabular}{lccp{16em}cp{6em}r}
            &   & $E_S$ & $Q_S$ \\ \hline
            &   & 2.773 & \parbox{15em}{What is traveling under the bridge archway ?} \\
                                                                 \\            
Sampling & $\Delta$  & $E_T$ & $Q_G$                                               & $E_G$ & VQA top-1     & $E_G - E_T$  \\
\hline
Pretrain & $-1$  & 0.773 & is the water calm ?                                                                             & 0.231 & no       & -0.542 \\
w/o A    & $0$   & 2.773 & how many birds are there ?                                                                      & 5.091 & no       & 2.318  \\
         & $+1$  & 4.773 & where are the boats ?                                                                           & 3.979 & no       & -0.794 \\
\hline
Pretrain & $-1$  & 0.773 & is the boat in the water ?                                                                      & 0.049 & no       & -0.724 \\
         & $0$   & 2.773 & what is the weather like ?                                                                      & 4.522 & wet      & 1.749  \\
         & $+1$  & 4.773 & where is the bird ?                                                                             & 3.123 & no       & -1.651 \\
\hline
FT w/o A & $-1$  & 0.773 & is there a lot facing to turn turn ?                                                            & 0.204 & no       & -0.569 \\
         & $0$   & 2.773 & is there a lot facing to turn turn ?                                                            & 0.204 & no       & -2.569 \\
         & $+1$  & 4.773 & where do people turn turn turn turn turn turn turn turn turn turn turn turn turn turn turn turn & 2.527 & in water & -2.246 \\
\hline
FT       & $-1$  & 0.773 & is this a passenger train ?                                                                     & 0.126 & no       & 0.648  \\
         & $0$   & 2.773 & who is going on the truck ?                                                                     & 1.620 & birds    & -1.153 \\
         & $+1$  & 4.773 & who is going on the car ?                                                                       & 2.229 & birds    & -2.545
\end{tabular}
}

    \caption{Effects on rephrasing with VQA attention.
    Rows with ``w/o A'' stand for without attention, otherwise with attention.
    There are three visual questions (images $I$ are on the left,
    and source questions $Q_S$ are at the top of each table.
    Top-1 answers predicted by VQA are shown for reference.
    }
    
    \label{fig:example_att}
\end{figure}

Figure \ref{fig:example_att} shows rephrasing results with or without attention
for the same three visual questions.
Results of Sampling-FT without attention are not good and
the same words appear repeatedly.
These results show that our model with the VQA attention in the encoder
works well.

\begin{table}[t]
 \centering
 
 \caption{Entropy errors, similarity and diversity scores of rephrasing results
 with and without VQA attention. Rows with ``w/o A'' stands for without attention, 
 otherwise with attention.
 }

 \label{tab:attention_difference}
 
\scalebox{0.5}{
 \begin{tabular}{c|l|cccccc}
$\Delta = E_T-E_S$    & Sampling & $|E_T - E_G|$  & BLEU-4 & CIDEr  & METEOR & ROUGE-L & Diversity \\  
\hline
\multirow{4}{*}{\centering $-2.0$}
            & Pretrain w/o A & 1.30$\pm$1.25 & 6.39   & 37.76 & 13.78  & 41.26   & 2967      \\  
            & Pretrain       & 1.46$\pm$1.35 & 8.73   & 57.19 & 15.61  & 45.47   & 2229      \\  
            & FT w/o A       & 0.76$\pm$0.65 & 0.08   & 1.56  & 5.39   & 20.26   & 1232      \\  
            & FT             & 0.91$\pm$0.83 & 0.74   & 5.38  & 6.87   & 27.96   & 1087      \\  
\hline
\multirow{4}{*}{\centering $-1.5$}
            & Pretrain w/o A & 1.33$\pm$1.22 & 7.73   & 45.61 & 14.24  & 42.09   & 3418      \\  
            & Pretrain       & 1.40$\pm$1.27 & 9.56   & 61.63 & 15.75  & 45.55   & 2361      \\  
            & FT w/o A       & 0.82$\pm$0.70 & 0.09   & 1.68  & 5.28   & 19.91   & 1523      \\  
            & FT             & 0.91$\pm$0.82 & 0.84   & 5.59  & 6.92   & 28.00   & 1217      \\  
\hline
\multirow{4}{*}{\centering $-1.0$}
            & Pretrain w/o A & 1.33$\pm$1.18 & 9.25   & 55.42 & 14.89  & 43.13   & 3759      \\  
            & Pretrain       & 1.38$\pm$1.22 & 11.19  & 72.19 & 16.37  & 46.51   & 2524      \\  
            & FT w/o A       & 0.89$\pm$0.76 & 0.07   & 1.74  & 5.17   & 19.54   & 1768      \\  
            & FT             & 0.93$\pm$0.83 & 0.80   & 5.46  & 6.87   & 27.73   & 1274      \\  
\hline
\multirow{4}{*}{\centering $-0.5$}
            & Pretrain w/o A & 1.14$\pm$1.12 & 10.09  & 62.00 & 15.36  & 43.94   & 4427      \\  
            & Pretrain       & 1.18$\pm$1.16 & 12.14  & 79.20 & 16.86  & 47.36   & 3117      \\  
            & FT w/o A       & 0.84$\pm$0.82 & 0.20   & 2.58  & 5.42   & 20.64   & 2097      \\  
            & FT             & 0.89$\pm$0.88 & 1.06   & 6.52  & 7.02   & 28.10   & 1550      \\  
\hline
\multirow{4}{*}{\centering $\pm 0.0$}
            & Pretrain w/o A & 0.95$\pm$1.05 & 10.43  & 65.89 & 15.49  & 44.31   & 5443      \\  
            & Pretrain       & 0.98$\pm$1.09 & 12.88  & 86.64 & 17.25  & 48.14   & 3734      \\  
            & FT w/o A       & 0.78$\pm$0.83 & 0.29   & 2.94  & 5.76   & 21.98   & 2518      \\  
            & FT             & 0.84$\pm$0.91 & 1.39   & 7.50  & 7.30   & 28.79   & 1846      \\  
\hline
\multirow{4}{*}{\centering $+0.5$}
            & Pretrain w/o A & 1.19$\pm$1.14 & 9.28   & 56.96 & 14.94  & 42.77   & 5049      \\  
            & Pretrain       & 1.26$\pm$1.20 & 11.00  & 71.25 & 16.27  & 45.85   & 3312      \\  
            & FT w/o A       & 1.02$\pm$0.91 & 0.24   & 2.50  & 5.45   & 20.76   & 2717      \\  
            & FT             & 0.95$\pm$0.86 & 1.37   & 7.25  & 7.59   & 29.62   & 1846      \\  
\hline
\multirow{4}{*}{\centering $+1.0$}
            & Pretrain w/o A & 1.53$\pm$1.18 & 8.56   & 51.10 & 14.53  & 41.75   & 2902      \\  
            & Pretrain       & 1.59$\pm$1.24 & 9.95   & 62.26 & 15.67  & 44.56   & 1874      \\  
            & FT w/o A       & 1.31$\pm$0.96 & 0.20   & 2.00  & 5.10   & 19.36   & 2790      \\  
            & FT             & 1.18$\pm$0.82 & 1.29   & 6.65  & 7.82   & 30.15   & 1712      \\  
\hline
\multirow{4}{*}{\centering $+1.5$}
            & Pretrain w/o A & 1.60$\pm$1.13 & 7.45   & 44.52 & 14.09  & 41.05   & 2937      \\  
            & Pretrain       & 1.68$\pm$1.16 & 8.77   & 54.53 & 15.24  & 43.72   & 1853      \\  
            & FT w/o A       & 1.55$\pm$1.07 & 0.14   & 1.52  & 4.77   & 17.99   & 2663      \\  
            & FT             & 1.36$\pm$0.85 & 1.19   & 6.03  & 7.86   & 30.22   & 1501      \\  
\hline
\multirow{4}{*}{\centering $+2.0$}
            & Pretrain w/o A & 1.61$\pm$1.04 & 6.57   & 40.14 & 13.82  & 40.70   & 2835      \\  
            & Pretrain       & 1.70$\pm$1.06 & 7.70   & 48.45 & 14.91  & 43.05   & 1718      \\  
            & FT w/o A       & 1.78$\pm$1.21 & 0.11   & 1.18  & 4.56   & 16.91   & 2435      \\  
            & FT             & 1.46$\pm$0.93 & 1.00   & 5.28  & 7.67   & 29.85   & 1345      \\  
\hline
\end{tabular}
}
\end{table}

\subsection{Weight $\lambda$ for entropy loss}

\begin{figure}[t]
    \centering
    \includegraphics[width=\linewidth]{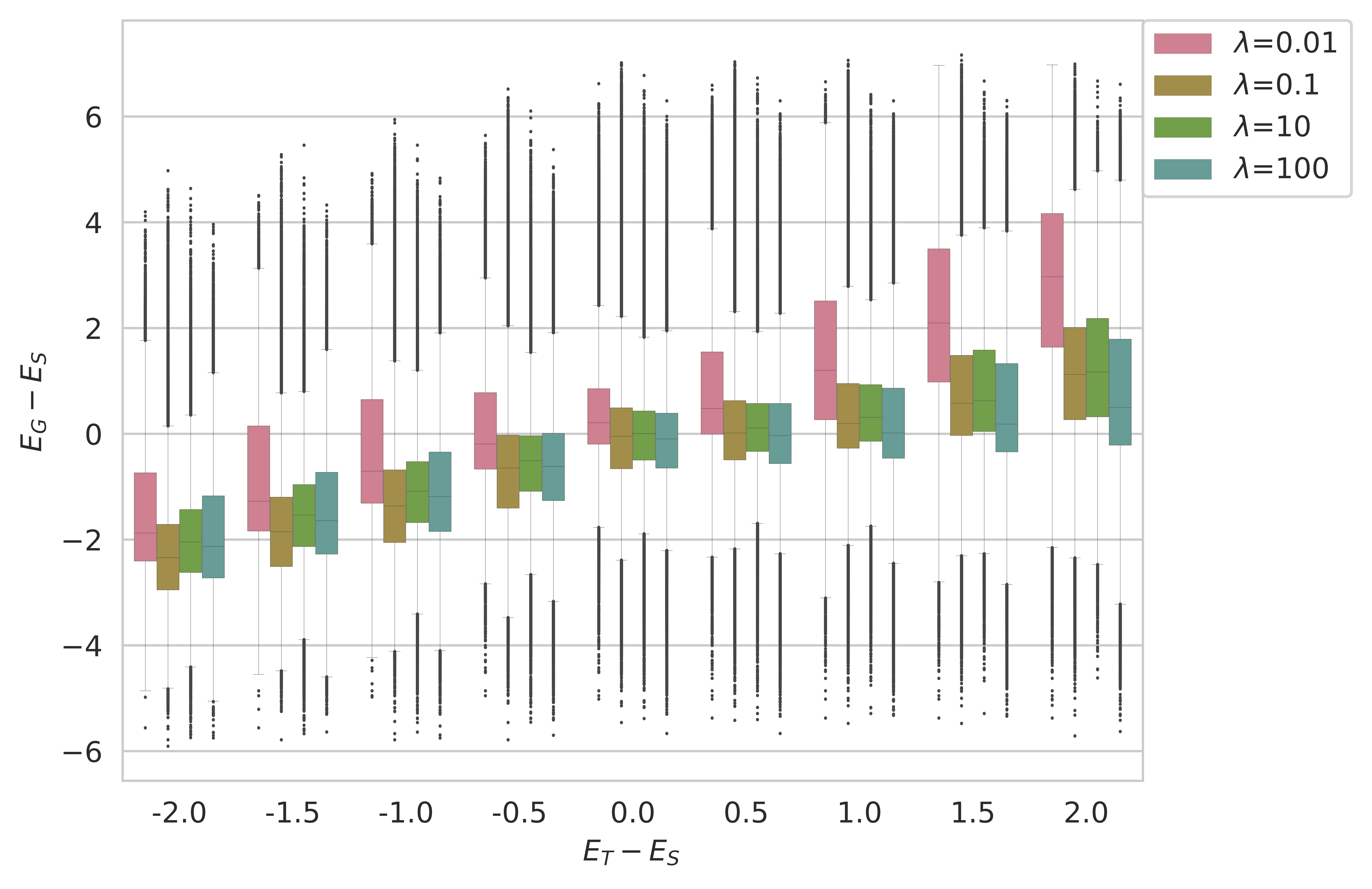}

    \caption{
    Effect of weight $\lambda$ on entropy errors for each value of $\Delta$ over all questions.
    The horizontal axis is the specified value of $\Delta = E_T - E_S$,
    and the vertical axis is the entropy error $E_G - E_T$ (without taking absolute).
     }

    \label{fig:boxplot_lambda}
\end{figure}

\begin{figure}[t]
\centering
 
\begin{minipage}{.2\linewidth}
\includegraphics[width=\linewidth]{VQG/example/image_1.jpg}
\end{minipage}
\scalebox{0.4}{
\begin{tabular}{lccp{16em}cp{6em}r}
            &   & $E_S$ & $Q_S$ \\ \hline
            &   & 2.899 & What is the word written on the field ? \\
                                                                 \\            
$\lambda$ & $\Delta$  & $E_T$ & $Q_G$                                               & $E_G$ & VQA top-1     & $E_G - E_T$  \\
\hline
0.01 & $-1$  & 0.899 & is the tennis ball in motion ?                                                                                                                  & 0.498 & no          & -0.401 \\
     & $0$   & 2.899 & what is the man doing ?                                                                                                                         & 5.392 & hit ball    & 2.493  \\
     & $+1$  & 4.899 & what is the man holding ?                                                                                                                       & 4.066 & nothing     & -0.833 \\
\hline
0.1  & $-1$  & 0.899 & are the players playing tennis ?                                                                                                                & 0.407 & no          & -0.492 \\
     & $0$   & 2.899 & how many people are in the picture ?                                                                                                            & 1.554 & 1           & -1.345 \\
     & $+1$  & 4.899 & what does the child say ?                                                                                                                       & 3.495 & serve       & -1.404 \\
\hline 
1     & $-1$  & 0.899 & is this a professional baseball game ?                     & 0.702 & no            & -0.197 \\
      & $0$   & 2.899 & who is going downhill ?                                    & 2.671 & tennis player & -0.228 \\
      & $+1$  & 4.899 & who is going on the bike ?                                 & 4.756 & man           & -0.143 \\
\hline 
10   & $-1$  & 0.899 & \unk \unk \unk \unk \unk \unk \unk \unk \unk \unk \unk \unk \unk \unk \unk \unk \unk \unk \unk \unk                                             & 1.875 & no          & 0.976  \\
     & $0$   & 2.899 & the license the license license license license license license license license license license license license license license license license & 0.519 & no          & -2.380 \\
     & $+1$  & 4.899 & where the license the license license license license license license license side side side side side side side side                           & 4.262 & bottom left & -0.637 \\
\hline
100  & $-1$  & 0.899 & sill swimsuits swimsuits \unk \unk \unk \unk \unk \unk \unk \unk \unk \unk \unk \unk \unk \unk \unk \unk                                        & 0.109 & no          & -0.79  \\
     & $0$   & 2.899 & who items items left left left left left left left left sculptures ? left sculptures ? left sculptures ?                                        & 4.045 & yes         & 1.146  \\
     & $+1$  & 4.899 & who items items left left left left left left left left left left left left left left left left                                                 & 5.38  & yes         & 0.481  
\end{tabular}
}

\vspace{.5cm}

\begin{minipage}{.2\linewidth}
\includegraphics[width=\linewidth]{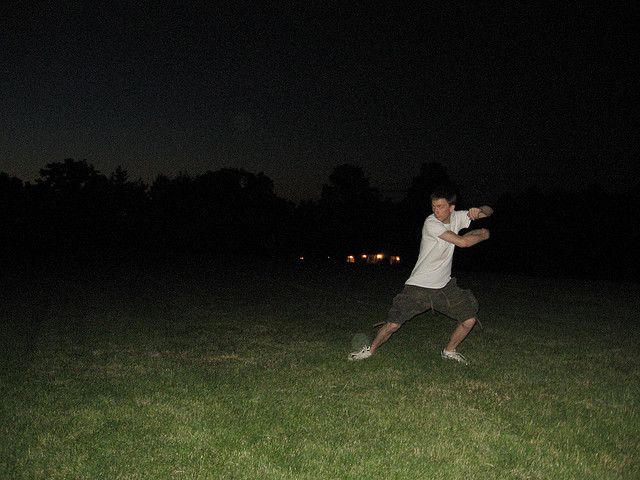}
\end{minipage}
\scalebox{0.4}{
\begin{tabular}{lccp{16em}cp{6em}r}
             &  & $E_S$ & $Q_S$ \\ \hline
             &  & 2.161 & What is he doing at night ? \\
                                                      \\            
$\lambda$ & $\Delta$  & $E_T$ & $Q_G$                                               & $E_G$ & VQA top-1     & $E_G - E_T$  \\
\hline
0.01 & $-1$  & 0.161 & is the man wearing a hat ?                                                                                           & 0.126 & no      & -0.035 \\
     & $0$   & 2.161 & is the man wearing a hat or a helmet ?                                                                               & 0.016 & no      & -2.145 \\
     & $+1$  & 4.161 & what is the man doing ?                                                                                              & 5.181 & nothing & 1.02   \\
\hline
0.1  & $-1$  & 0.161 & does the man have a beard ?                                                                                          & 0.025 & no      & -0.136 \\
     & $0$   & 2.161 & does the man have a beard ?                                                                                          & 0.025 & no      & -2.136 \\
     & $+1$  & 4.161 & what does the man have on his hand ?                                                                                 & 0.918 & nothing & -3.243 \\
\hline
1           & $-1$  & 0.161             & will the kite in the picture ?                         & 0.601 & no       & 0.441  \\
            & $0$   & 2.161             & which two people going downhill ?                      & 2.341 & man      & 0.181  \\
            & $+1$  & 4.161             & why does the person have two feet in the air ?         & 4.09  & playing frisbee & -0.07 \\
\hline
10   & $-1$  & 0.161 & is this license both side both side both side both both both both both both both both both both                      & 0.712 & yes     & 0.551  \\
     & $0$   & 2.161 & color the license both side both side both side both side both side both side both side both side                    & 2.301 & white   & 0.14   \\
     & $+1$  & 4.161 & where the license the license the license license license license license license side side side side side side side & 3.727 & nowhere & -0.433 \\
\hline
100  & $-1$  & 0.161 & \unk \unk \unk \unk \unk \unk \unk \unk \unk \unk \unk \unk \unk \unk \unk \unk \unk \unk \unk \unk                  & 2.223 & no      & 2.063  \\
     & $0$   & 2.161 & \unk \unk \unk \unk \unk \unk \unk \unk \unk \unk \unk \unk \unk \unk \unk \unk \unk \unk \unk \unk                  & 2.223 & no      & 0.063  \\
     & $+1$  & 4.161 & who items items left left left left left left left left left left left left left left left left                      & 4.197 & nothing & 0.036  
\end{tabular}
}

\vspace{.5cm}

\begin{minipage}{.2\linewidth}
\includegraphics[width=\linewidth]{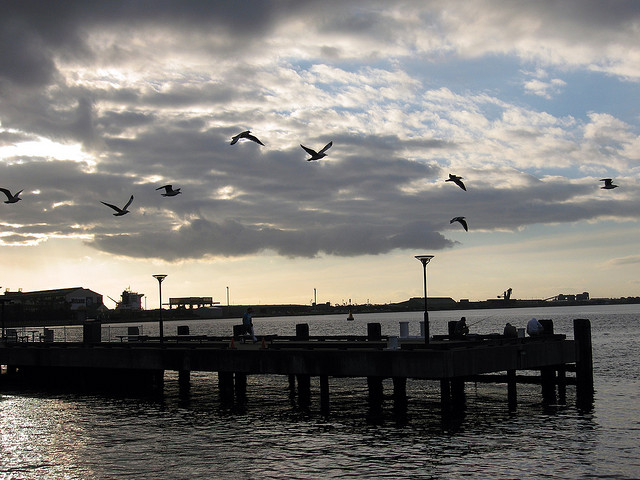}
\end{minipage}
\scalebox{0.4}{
\begin{tabular}{lccp{16em}cp{6em}r}
          &     & $E_S$ & $Q_S$ \\ \hline
          &     & 2.773 & \parbox{15em}{What is traveling under the bridge archway ?} \\
                                                                 \\            
$\lambda$ & $\Delta$  & $E_T$ & $Q_G$                                               & $E_G$ & VQA top-1     & $E_G - E_T$  \\
\hline
0.01 & $-1$  & 0.773 & is this a lake ?                                                                                                                                                                           & 0.074 & no         & -0.7   \\
     & $0$   & 2.773 & what is the boat tied to ?                                                                                                                                                                 & 6.229 & colgate    & 3.456  \\
     & $+1$  & 4.773 & what time is it ?                                                                                                                                                                          & 3.385 & evening    & -1.388 \\
\hline
0.1  & $-1$  & 0.773 & are there boats in the picture ?                                                                                                                                                           & 0.007 & no         & -0.767 \\
     & $0$   & 2.773 & how many boats are there ?                                                                                                                                                                 & 2.37  & no         & -0.404 \\
     & $+1$  & 4.773 & how can you tell the boat is in the boat \unk window ?                                                                                                                                     & 5.04  & reflection & 0.267  \\
\hline
1           & $-1$ & 0.773 & is this a passenger train ? & 0.126 & no    & 0.648  \\
            & $0$  & 2.773 & who is going on the truck ? & 1.62  & birds & -1.153 \\
            & $+1$ & 4.773 & who is going on the car ?   & 2.229 & birds & -2.545 \\
\hline
10   & $-1$  & 0.773 & color both both both both both both both both both both both both both both both both both both                                                                                            & 0.681 & black      & -0.092 \\
     & $0$   & 2.773 & color the license patterns patterns patterns patterns ?                                                                                                                                    & 0.845 & black      & -1.928 \\
     & $+1$  & 4.773 & where the license the license license license license license license license side side side side side side side side                                                                      & 3.052 & unknown    & -1.721 \\
\hline
100  & $-1$  & 0.773 & who items items sculptures sculptures items sculptures sculptures sculptures sculptures sculptures sculptures sculptures sculptures sculptures sculptures sculptures sculptures sculptures & 2.572 & no         & 1.799  \\
     & $0$   & 2.773 & who items items left left left left left left left left left left left left left left sculptures items                                                                                     & 0.59  & birds      & -2.183 \\
     & $+1$  & 4.773 & who items items left left left left left left left left left left left left left left left left                                                                                            & 0.637 & birds      & -4.137 
\end{tabular}
}

    \caption{Effects on rephrasing for different values of the weight $\lambda$.
    There are three visual questions (images $I$ are on the left,
    and source questions $Q_S$ are at the top of each table).
    Top-1 answers predicted by VQA are shown for reference.
    }
    
    \label{fig:example_lambda}
\end{figure}

For all experiments above
the weight $\lambda$ of the entropy loss was $\lambda=1$,
but this weight should have a large impact on the result
because it trades off between a simple VQG model and our entropy-aware rephrasing model.
Table \ref{tab:lambda_difference} shows results
for Sampling-FT with different values of $\lambda$.
Results are shown in Table \ref{tab:lambda_difference},
Figure \ref{fig:boxplot_lambda}, and Figure \ref{fig:example_lambda}.

The entropy errors become small when $\lambda=10$ or 1.
$\lambda=100$ is too large because the entropy error increases.
However similarity scores for $\lambda=10$ and 100 are quite low
because grammatically correct question sentences are not generated (as shown in Fig. \ref{fig:example_lambda})
and training becomes unstable.
The case of $\lambda=0.01$ has higher similarity scores 
because it is close to the Sampling Pretrain sisutation,
and questions similar to source questions are generated.

\begin{table}[t]
 \centering

 \caption{Entropy errors, similarity and diversity scores of rephrasing results
 for different values of weight $\lambda$. Rows with ``w/o A'' stands for without attention, 
 otherwise with attention.
 }

 \label{tab:lambda_difference}
 
\scalebox{0.5}{
 \begin{tabular}{c|l|cccccc}
$\Delta = E_T-E_S$    & $\lambda$ & $|E_T - E_G|$  & BLEU-4 & CIDEr  & METEOR & ROUGE-L & Diversity \\  
\hline
\multirow{5}{*}{\centering $-2.0$}
            & 0.01 & 1.05$\pm$1.08 & 5.72   & 38.36 & 13.64  & 40.04   & 1966      \\  
            & 0.1  & 0.92$\pm$0.82 & 2.36   & 18.30 & 9.83   & 31.08   & 2084      \\  
            & 1.0  & 0.91$\pm$0.83 & 0.74   & 5.38  & 6.87   & 27.96   & 1087      \\  
            & 10   & 0.82$\pm$0.76 & 0.00   & 0.18  & 2.48   & 10.10   & 2561      \\  
            & 100  & 1.01$\pm$0.89 & 0.00   & 0.05  & 0.43   & 2.32    & 3148      \\  
\hline
\multirow{5}{*}{\centering $-1.5$}
            & 0.01 & 1.12$\pm$1.09 & 6.86   & 44.82 & 13.97  & 40.52   & 2275      \\  
            & 0.1  & 0.96$\pm$0.83 & 2.71   & 20.77 & 9.98   & 31.72   & 2508      \\  
            & 1.0  & 0.91$\pm$0.82 & 0.84   & 5.59  & 6.92   & 28.00   & 1217      \\  
            & 10   & 0.81$\pm$0.73 & 0.00   & 0.15  & 2.48   & 9.81    & 2725      \\  
            & 100  & 1.01$\pm$0.86 & 0.00   & 0.06  & 0.43   & 2.36    & 3740      \\  
\hline
\multirow{5}{*}{\centering $-1.0$}
            & 0.01 & 1.12$\pm$1.05 & 7.93   & 52.23 & 14.35  & 41.18   & 2485      \\  
            & 0.1  & 0.99$\pm$0.84 & 2.94   & 21.51 & 10.08  & 32.08   & 2989      \\  
            & 1.0  & 0.93$\pm$0.83 & 0.80   & 5.46  & 6.87   & 27.73   & 1274      \\  
            & 10   & 0.79$\pm$0.73 & 0.00   & 0.16  & 2.49   & 9.49    & 2921      \\  
            & 100  & 1.00$\pm$0.84 & 0.00   & 0.06  & 0.42   & 2.37    & 4263      \\  
\hline
\multirow{5}{*}{\centering $-0.5$}
            & 0.01 & 0.99$\pm$1.00 & 8.57   & 56.90 & 14.88  & 42.23   & 2839      \\  
            & 0.1  & 0.94$\pm$0.88 & 3.10   & 21.48 & 10.30  & 32.12   & 3738      \\  
            & 1.0  & 0.89$\pm$0.88 & 1.06   & 6.52  & 7.02   & 28.10   & 1550      \\  
            & 10   & 0.71$\pm$0.69 & 0.00   & 0.17  & 2.75   & 9.89    & 3234      \\  
            & 100  & 0.89$\pm$0.83 & 0.00   & 0.10  & 0.47   & 2.97    & 5257      \\  
\hline
\multirow{5}{*}{\centering $\pm 0.0$}
            & 0.01 & 0.84$\pm$0.94 & 8.93   & 59.36 & 15.16  & 42.85   & 3302      \\  
            & 0.1  & 0.86$\pm$0.89 & 3.27   & 21.41 & 10.54  & 31.90   & 4701      \\  
            & 1.0  & 0.84$\pm$0.91 & 1.39   & 7.50  & 7.30   & 28.79   & 1846      \\  
            & 10   & 0.65$\pm$0.67 & 0.00   & 0.17  & 2.97   & 10.41   & 3886      \\  
            & 100  & 0.80$\pm$0.82 & 0.00   & 0.13  & 0.51   & 3.44    & 6652      \\  
\hline
\multirow{5}{*}{\centering $+0.5$}
            & 0.01 & 0.97$\pm$0.94 & 9.69   & 64.00 & 15.73  & 43.75   & 3229      \\  
            & 0.1  & 0.95$\pm$0.83 & 4.02   & 24.85 & 11.07  & 32.98   & 4959      \\  
            & 1.0  & 0.95$\pm$0.86 & 1.37   & 7.25  & 7.59   & 29.62   & 1846      \\  
            & 10   & 0.75$\pm$0.67 & 0.00   & 0.18  & 2.98   & 10.24   & 3866      \\  
            & 100  & 1.00$\pm$0.82 & 0.00   & 0.11  & 0.46   & 2.90    & 7156      \\  
\hline
\multirow{5}{*}{\centering $+1.0$}
            & 0.01 & 1.23$\pm$0.96 & 9.34   & 61.11 & 15.46  & 42.96   & 3099      \\  
            & 0.1  & 1.11$\pm$0.79 & 4.79   & 28.42 & 11.59  & 34.10   & 5321      \\  
            & 1.0  & 1.18$\pm$0.82 & 1.29   & 6.65  & 7.82   & 30.15   & 1712      \\  
            & 10   & 0.98$\pm$0.67 & 0.00   & 0.18  & 2.90   & 9.91    & 3662      \\  
            & 100  & 1.26$\pm$0.84 & 0.00   & 0.08  & 0.41   & 2.40    & 7681      \\  
\hline
\multirow{5}{*}{\centering $+1.5$}
            & 0.01 & 1.43$\pm$1.04 & 8.62   & 55.33 & 15.03  & 42.25   & 2399      \\  
            & 0.1  & 1.23$\pm$0.83 & 5.27   & 29.50 & 11.86  & 34.97   & 5312      \\  
            & 1.0  & 1.36$\pm$0.85 & 1.19   & 6.03  & 7.86   & 30.22   & 1501      \\  
            & 10   & 1.15$\pm$0.74 & 0.00   & 0.17  & 2.74   & 9.41    & 3316      \\  
            & 100  & 1.49$\pm$0.92 & 0.00   & 0.07  & 0.36   & 1.92    & 7797      \\  
\hline
\multirow{5}{*}{\centering $+2.0$}
            & 0.01 & 1.55$\pm$1.02 & 7.73   & 48.18 & 14.76  & 42.43   & 1879      \\  
            & 0.1  & 1.32$\pm$0.92 & 5.14   & 27.44 & 11.78  & 35.12   & 5060      \\  
            & 1.0  & 1.46$\pm$0.93 & 1.00   & 5.28  & 7.67   & 29.85   & 1345      \\  
            & 10   & 1.26$\pm$0.83 & 0.00   & 0.15  & 2.53   & 8.67    & 2837      \\  
            & 100  & 1.69$\pm$1.05 & 0.00   & 0.05  & 0.31   & 1.48    & 7865      \\  
\hline
\end{tabular}
}
\end{table}

\section{Conclusion}

In this paper, we have proposed a task of
rephrasing visual questions so that the rephrased question has the specified ambiguity in terms of entropy.
Our proposed model has an encoder-decoder architecture,
followed by a pre-train VQA model for entropy computation.
Experimental results when using the proposed
learning strategies with the VQA v2 dataset
have shown that the model can minimize
the error to the specified entropy value.

One of the limitations of our work is that
rephrased questions are often irrelevant to the given image
while entropy is close to the specified value.
To tackle this limitation,
it may be necessary to explore architectures for 
extracting the concept of the image
in addition to VQA attention
and applying it to the rephrased question.
Another limitation is the small diversity of the
rephrased questions. To avoid the generation of similar questions,
the variational approach \cite{krishna2019information}
might be one of the promising solutions.
And last but no least, the construction of datasets specific
to this task is the most challenging part of this work
and we will continue to explore how to define
the ambiguity of visual questions
for dataset annotations.

\section*{Acknowledgement}

We would like to thank 
Yoshitaka Ushiku for his comments as a mentor of the PRMU mentorship program.
This work was supported by 
JSPS KAKENHI grant number JP16H06540.

\bibliographystyle{ieee_fullname}
\bibliography{reference}

\end{document}